\documentclass[11pt]{article}
\usepackage[preprint]{acl}   
\usepackage{times}
\usepackage{latexsym}
\usepackage{url}
\usepackage[T1]{fontenc}
\usepackage[utf8]{inputenc}

\usepackage{amsmath}
\usepackage{graphicx}
\usepackage{booktabs}

\newcommand{\rhosym}{\ensuremath{\rho}}
\newcommand{\bielik}{Bielik}
\newcommand{\gemma}{Gemma-4}
\newcommand{\qwen}{Qwen3}
\newcommand{\pllum}{PLLuM}

\title{Graded Entity-Familiarity Readouts in Language Models:\\
Polish Adaptation, Cross-Language Robustness, and Refusal Steering\thanks{Preprint. Under review.}}

\author{
  Grzegorz Brzezinka \\
  Prosit AS \\
  \texttt{greg@prosit.no}
}

\begin{document}

\maketitle

\begin{abstract}
Can a language model estimate its familiarity with an entity before generating an answer? We study activations at the final prompt token in twelve instruction-tuned models from the \bielik{}, \pllum{}, \gemma{}, and \qwen{} families, using a new dataset of 1{,}440 Polish entities spanning four domains and ten Wikipedia-pageview deciles, plus fabricated controls. Familiarity-probe scores separate real from fabricated entities in every family; in the Polish-adapted \bielik{} and \pllum{} families they additionally track entity popularity (model-mean Spearman \rhosym{} 0.28--0.57, versus at most 0.11 in \gemma{} and \qwen{}), a pattern more strongly associated with Polish adaptation than with parameter count in this model sample. In a paired experiment on two families, probes retain 96--101\% of within-language AUROC when the Polish question stem is replaced with an English one around unchanged entity names, showing robustness to prompt language in this setting. In \gemma{}-12B, the only model that natively refuses, adding a one-dimensional familiarity direction at a single layer moves refusal rates monotonically in both directions (0.24 to 1.00 on well-known entities; 0.73 to 0.00 on unknown ones). Finally, a calibrated familiarity probe is competitive among pre-generation abstention gates, although post-generation detectors better predict behavioral error on average. These results support a graded pre-generation entity-familiarity readout, and a separation between representational familiarity and the policy that converts it into abstention.
\end{abstract}

\section{Introduction}
\label{sec:intro}

When a language model is asked about an entity it has never encountered, it can either abstain or confabulate. Prior work shows that the model's internal states contain information relevant to this choice \citep{kadavath2022languagemodels,azaria2023internal,orgad2025llmsknow}, and \citet{ferrando2025entity} localized an entity-recognition mechanism in sparse-autoencoder space. Closest to our own work, \citet{gottesman2024keen} showed that supervised probes over an entity's \emph{pre-generation} hidden states predict graded QA accuracy. Most detection methods, however, score \emph{generated answers} \citep{ding2025d2hscore,ettori2025eigentrack,su2024mind,farquhar2024detecting}. A recent study of the Polish \bielik{} family \citep{brzezinka2026dispersion} instead measured the \emph{prompt point} (the last token of the question, before any answer token exists) and found that unsupervised activation-dispersion metrics separate well-known from fabricated Polish entities at AUROC 0.95--1.00 across scales from 1.5B to 11B parameters, while the same signal says much less about whether the answer will be factually correct.

That result was binary (known vs.\ fabricated), single-family, single-language, and purely correlational. It therefore left open the questions that determine what the signal \emph{is}:

\begin{itemize}
\item \textbf{RQ1 (language).} Is the signal about the entity or the question's surface form? If a probe trained on Polish questions transfers to English questions about the same entities, the readout is at least robust to the question-stem language.
\item \textbf{RQ2 (gradation).} Is familiarity a bit or a quantity? If the score tracks a continuous popularity axis, it behaves as a graded quantity, and the predecessor's ceiling is a property of an easy contrast, not of the signal.
\item \textbf{RQ3 (causality).} Does the model itself read this signal, or is the familiarity--refusal link merely correlational?
\item \textbf{RQ4 (application).} Can the signal gate generation (abstain or route to retrieval \emph{before} producing an answer) competitively with post-generation detectors, at a fraction of their cost?
\end{itemize}

We answer all four with one fixed one-forward-pass measurement protocol (\S\ref{sec:signals}), applied to a new popularity-graded dataset of Polish entities (\S\ref{sec:dataset}) and twelve instruction-tuned models from four families spanning a Polish-pretraining axis, from Polish-native \bielik{} through \pllum{} (Polish continual pretraining on Llama-3.1 and Mistral-NeMo bases) to the multilingual, non-Polish-centric \gemma{} and \qwen{}. Coverage varies by question: gradation (RQ2) uses all twelve models on the new dataset (v2) and gating (RQ4) the largest model per family on v2, while language transfer (RQ1; six models) and steering (RQ3; \gemma{}-12B) predate v2 and use the predecessor's three-condition dataset (v1; \S\ref{sec:dataset}).

Our contributions, in narrative order:
(1)~a QID-resolved, decile-stratified Polish entity dataset with fabricated anchors, token-length matched to the real distribution (\S\ref{sec:dataset});
(2)~evidence that the familiarity score is graded in the Polish-adapted families and that its gradation is more strongly associated with Polish adaptation than with scale in this model sample, with a behavioral mirror in answer correctness (\S\ref{sec:gradation});
(3)~zero-shot cross-language probe transfer at 96--101\% of within-language performance in the two families tested, arguing against a large question-stem-language effect in this paired setting (\S\ref{sec:language}); and
(4)~an intervention-plus-application study: a single rank-one familiarity direction steers \gemma{}-12B refusal monotonically in both directions (0.24${\to}$1.00 / 0.73${\to}$0.00), extending \citet{ferrando2025entity} with a dose--response characterization (\S\ref{sec:causal}), and a risk--coverage comparison against adapted post-generation baselines that positions the calibrated probe within the pre-generation gate class and quantifies the dissociation between representational familiarity and behavioral abstention policy (\S\ref{sec:gating}, \S\ref{sec:discussion}).

\section{Related Work}
\label{sec:related}

\paragraph{Internal-state hallucination detection.}
Supervised probes on hidden states detect falsehoods and hallucinations \citep{azaria2023internal,orgad2025llmsknow,du2024haloscope,chen2024inside,sriramanan2024llmcheck,binkowski2025lapeigvals,park2025tsv}; semantic entropy measures answer instability across samples \citep{kuhn2023semantic,farquhar2024detecting,kossen2024semantic}; SelfCheckGPT is sampling-based and black-box \citep{manakul2023selfcheckgpt}; unsupervised linear truth directions \citep{burns2023dlk,marks2024geometry} and verbalized confidence \citep{lin2022teaching,tian2023just} offer complementary readouts. Closest to our setting are question-only approaches: \citet{kadavath2022languagemodels} elicit self-knowledge behaviorally, \citet{cencerrado2025noanswer} train question-side probes, \citet{slobodkin2023curious} find answerability encoded before generation, \citet{snyder2024early} detect hallucinations from pre-generation artifacts, and \citet{ferrando2025entity} identify entity-recognition features that causally modulate refusal in Gemma~2. \citet{gottesman2024keen} estimate entity knowledge from pre-generation hidden states with supervised probes that predict graded QA accuracy; this is the nearest precedent to our readout. We differ in the popularity-decile design with a fabricated anchor and in adding cross-family, cross-language, and interventional axes; of our two readouts, only dispersion is unsupervised and closed-form, while our best-performing probe is, like theirs, a supervised classifier. \citet{obeso2025realtime} carry entity-recognition probes into real-time detection. Linear representations are known to track pretraining term frequency \citep{merullo2025linear,mozafari2026pretraining}, and internal states reflect knowledge recall rather than truthfulness \citep{cheang2026internalstates}, converging with our familiarity-not-truth framing. Recent single-generation detectors such as D\textsuperscript{2}HScore \citep{ding2025d2hscore}, EigenTrack \citep{ettori2025eigentrack}, and MIND \citep{su2024mind} score the answer's hidden states; we compare against documented adaptations of all three (\S\ref{sec:gating}).

\paragraph{Popularity and the long tail.}
Factual accuracy tracks entity popularity \citep{mallen2023trust} and pretraining co-occurrence \citep{kandpal2023longtail}. We turn this external axis into a probe of the \emph{internal} signal: if familiarity is a readout of exposure, its score should be monotone in log-popularity.

\paragraph{Activation steering.}
Adding difference-of-means directions to the residual stream steers behavior \citep{turner2023activation,rimsky2024steering}; inference-time intervention steers truth-related directions \citep{li2023iti}; refusal in particular is mediated by a single direction \citep{arditi2024refusal}, and conditional or surgical variants refine where and when to steer \citep{lee2025cast,yang2025casal}. We apply this methodology not to jailbreaking but to the epistemic axis: does moving along the familiarity direction move the model's own abstention decision?

\paragraph{Polish evaluation.}
Mu-SHROOM \citep{vazquez2025mushroom} and MultiHaluDet \citep{alvi2026multihaludet} cover Polish hallucination annotation; \bielik{} \citep{ociepa2025bielikv3small,ociepa2025bielik11bv2} and \pllum{} \citep{kocon2025pllum} are the Polish-adapted model families we study, alongside \gemma{} \citep{gemmateam2026gemma4} and \qwen{} \citep{yang2025qwen3}; \pllum{}'s 8B and 12B variants build on Llama-3.1 \citep{grattafiori2024llama3} and Mistral-NeMo \citep{mistral2024nemo}.

\section{Familiarity Readouts}
\label{sec:signals}

All measurements are taken at the \emph{prompt point}: the final token of a one-sentence Polish question (\emph{Kim jest \{entity\}? Odpowiedz jednym zdaniem.} for people; \emph{Czym jest \ldots} for places), in one forward pass, before generation. Following \citet{brzezinka2026dispersion} we use two signal classes plus a baseline: \textbf{dispersion} (unsupervised: per-layer inverse participation ratio and Shannon entropy over the post-SwiGLU MLP activation vector; best (metric, layer) cell selected on an anchor contrast, here top-decile-vs-fabricated, oriented so higher = more familiar), a \textbf{probe} (logistic regression on the residual-stream hidden state, layer selected by 5-fold CV, training-contrast scores always out-of-fold), and \textbf{first-token entropy}. Conventions: separability $=\max(\text{AUROC}, 1-\text{AUROC})$; bootstrap 95\% CIs; fixed seeds. Functionally analogous readouts appear in every family (on \gemma{}-12B the best layers sit in the same mid-depth relative band as \bielik{}, 0.64--0.71 on athletes, and the per-head picture is equally diffuse: rank consistency 0.57 vs.\ 0.56, 2 of 768 heads in all four domain top-20 sets), so one measurement recipe applies to every family without retuning. Because each readout is trained or selected per model, similar performance and layer placement do not establish that the families share an identical representation; cross-model transfer or representational-similarity tests would be needed for that stronger claim.

\section{A Popularity-Graded Dataset}
\label{sec:dataset}

\paragraph{Why graded.}
The predecessor's known-vs-fabricated contrast is at ceiling already at 1.5B, capping what scale comparisons can reveal. A pilot on its 336 real entities found the familiarity score is not a bit but a \emph{graded popularity readout}: Spearman correlation with log 12-month pl.wiki pageviews averaged $\rhosym=0.75$ across 6 models $\times$ 4 domains, surviving within the well-known tier alone; pageviews and Wikidata sitelinks were interchangeable. This motivated sampling the popularity axis continuously.

\paragraph{Design.}
For each of four domains (athletes, cities, writers, musicians) we harvest a full candidate pool from Wikidata SPARQL (occupation- or class-filtered, Polish citizenship or location, Polish-Wikipedia article required), resolve every entity by QID (excluding disambiguation pages by construction) and stratify 300 real entities into ten deciles of log 12-month pl.wiki pageviews (30 per decile), plus 60 fabricated anchor entities per domain: morphologically valid invented Polish names, screened for non-existence by exact page lookup and full-text search, and selected so that real-vs-fabricated token-count AUROC stays near chance under both the smallest and largest \bielik{} tokenizers (achieved 0.50--0.57, target $<0.65$); a post-hoc check confirms the screen generalizes to all four family tokenizers (pooled token-count AUROC 0.49--0.52, worst single domain cell 0.56). All sampling is deterministic (\texttt{random.Random(0)}); the totals are 1{,}200 real + 240 fabricated = 1{,}440 entities. The steering and language-transfer experiments (\S\ref{sec:language}, \S\ref{sec:causal}) predate v2 and use the predecessor's three-condition dataset (42 well-known / 42 obscure-real / 42 fabricated entities per domain), which we call v1.

\paragraph{Models.}
Twelve instruction-tuned checkpoints: \bielik{} v3.0 (1.5B, 4.5B, Minitron-7B \citep{kinas2026minitron}, 11B), \gemma{} (E4B, 12B), \pllum{} (4B; Llama-\pllum{}-8B; 12B, Mistral-NeMo-based; all instruct-2512), and \qwen{} (1.7B, 4B, 14B, with thinking disabled). The families span the axis we care about: Polish-native pretraining (\bielik{}), Polish continual pretraining over non-Polish bases (\pllum{}), and multilingual non-Polish-centric training (\gemma{}, \qwen{}).

\section{RQ2: Familiarity Is Graded and Aligned with Polish Adaptation}
\label{sec:gradation}

\begin{figure*}[t]
\centering
\includegraphics[width=\textwidth]{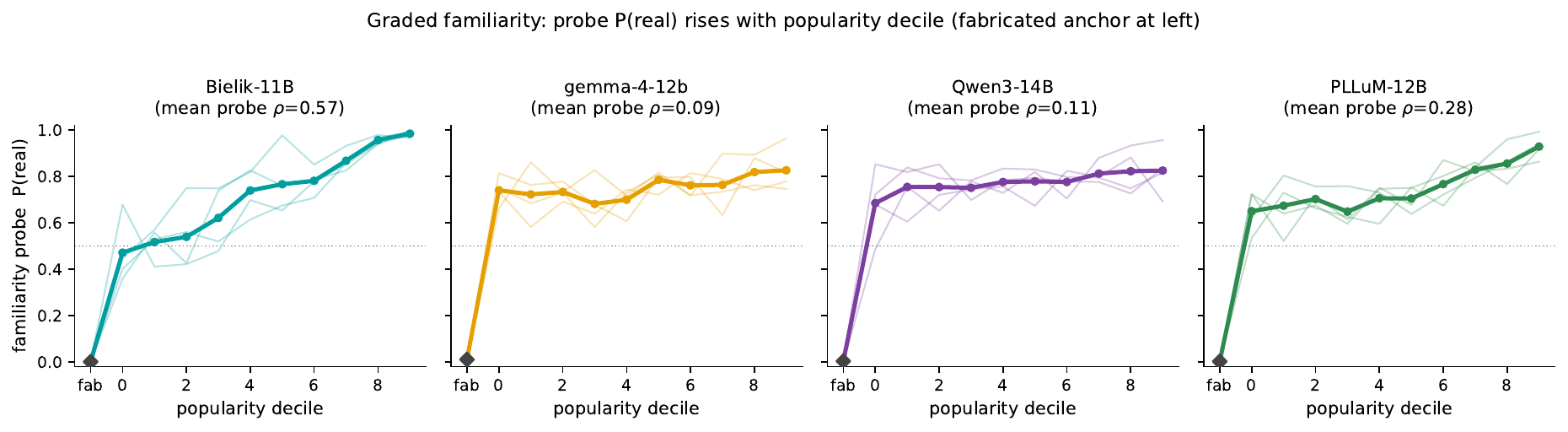}
\caption{Per-decile familiarity curves on dataset v2 (largest model per family; probe $P(\text{real})$; fabricated anchor at left, then popularity deciles 0--9; thin lines are individual domains). \bielik{}-11B rises through the entire popularity range (mean $\rhosym=0.57$); \gemma{}-12B and \qwen{}-14B separate the fabricated anchor from real entities but vary little across popularity deciles ($\rhosym=0.09/0.11$); Polish-adapted \pllum{}-12B is intermediate ($\rhosym=0.28$).}
\label{fig:gradation}
\end{figure*}

\begin{figure*}[t]
\centering
\includegraphics[width=0.92\textwidth]{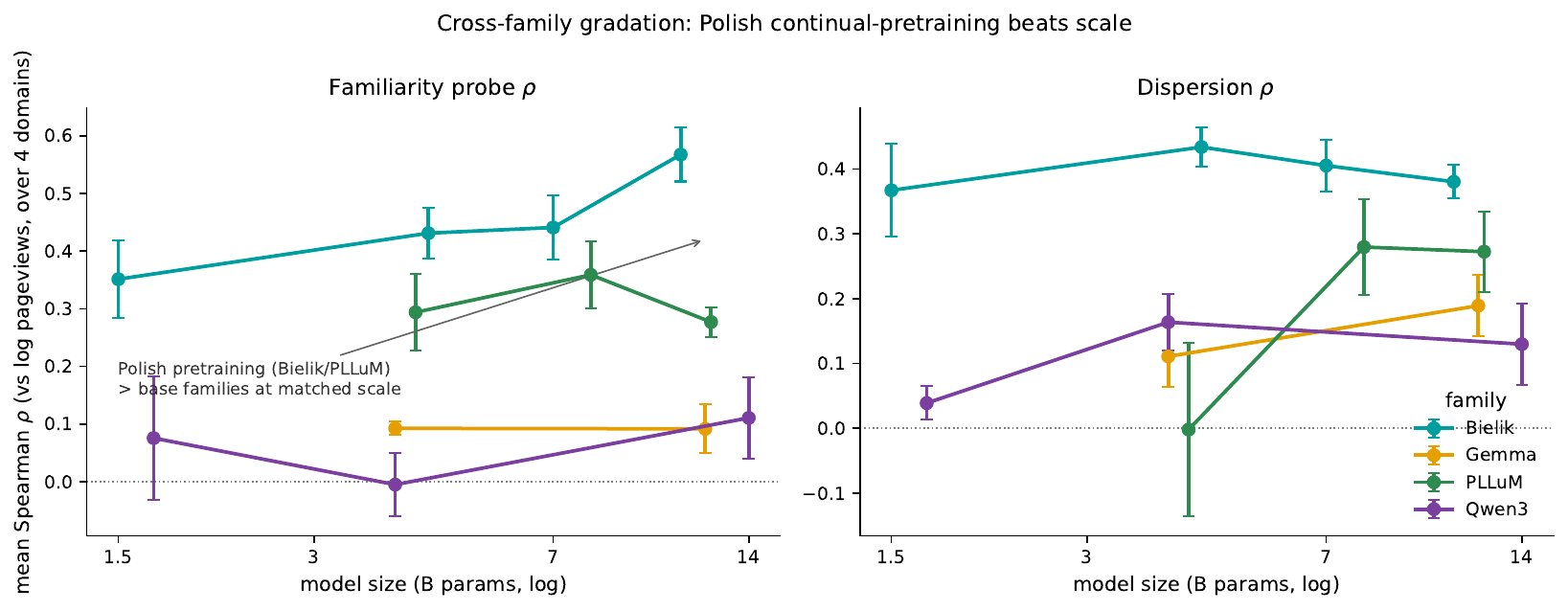}
\caption{Gradation strength (Spearman \rhosym{} of familiarity score vs.\ log pageviews among real entities, mean over four domains) by model size and family. \qwen{} stays near zero from 1.7B to 14B, while the Polish-adapted models (\bielik{}, \pllum{}) show stronger gradation at every matched size.}
\label{fig:familyrho}
\end{figure*}

\begin{table}[t]
\centering
\footnotesize
\setlength{\tabcolsep}{4pt}
\begin{tabular}{llrrr}
\toprule
Family & Model & B & disp.\ \rhosym & probe \rhosym \\
\midrule
\bielik & 1.5B & 1.5 & 0.367 & 0.351 \\
\bielik & 4.5B & 4.5 & 0.434 & 0.431 \\
\bielik & Minitron-7B & 7.0 & 0.405 & 0.441 \\
\bielik & 11B & 11.0 & 0.381 & \textbf{0.567} \\
\midrule
\gemma & E4B & 4.0 & 0.111 & 0.093 \\
\gemma & 12B & 12.0 & 0.189 & 0.092 \\
\midrule
\pllum & 4B & 4.3 & $-$0.002 & 0.294 \\
\pllum & Llama-8B & 8.0 & 0.280 & 0.359 \\
\pllum & 12B & 12.25 & 0.273 & 0.277 \\
\midrule
\qwen & 1.7B & 1.7 & 0.039 & 0.076 \\
\qwen & 4B & 4.0 & 0.164 & $-$0.005 \\
\qwen & 14B & 14.0 & 0.130 & 0.111 \\
\bottomrule
\end{tabular}
\caption{Gradation (Spearman \rhosym{} of familiarity score vs.\ $\log_{10}(\text{pageviews}+1)$ among real entities; mean over four domains; bootstrap CIs per cell in the released artifacts). Dispersion = best-layer metric at the top-decile-vs-fabricated anchor; probe = $P(\text{real})$ from a top-3-deciles-vs-fabricated logistic probe. Caveat: dispersion's cell selection uses the anchor AUROC, which couples it to $\rhosym$ (Spearman 0.795 across the 48 cells), a mild optimistic bias; the probe is exempt, as its scores are out-of-fold.}
\label{tab:rho}
\end{table}

\paragraph{The signal is graded.}
Figure~\ref{fig:gradation} shows per-decile familiarity curves; Table~\ref{tab:rho} and Figure~\ref{fig:familyrho} summarize gradation strength. Classifying each of the 48 model$\times$domain decile-vs-fabricated AUROC curves by their trend (Spearman $\rho$ of the ten AUROCs vs.\ decile index: rising $\rho\ge0.5$, falling $\rho\le-0.5$, else flat), 27 of 48 are rising and none crosses the falling threshold. This trichotomy is coarse (ten deciles, an arbitrary $\rho$ cutoff), so we read it as a descriptive summary rather than primary evidence; the per-cell correlations with uncertainty are in the released artifacts.

\paragraph{Gradation aligns with Polish adaptation, not scale, in this sample.}
Within \bielik{}, the probe's mean gradation grows from $\rhosym=0.351$ (1.5B) to $0.567$ (11B). But scale alone does not produce gradation: \qwen{} moves from $0.076$ to $0.111$ between 1.7B and 14B, and \gemma{}-12B sits at $0.092$. The most informative observational comparison is \pllum{}: after continual pretraining on Polish text, the Llama-3.1-based 8B reaches $\rhosym=0.359$ and the Mistral-NeMo-based 12B $0.277$, exceeding the matched-size non-Polish models by 0.17--0.27 absolute (\gemma{}-12B $0.092$, \qwen{}-14B $0.111$). We now have controlled before/after evidence for both \pllum{} architectures: each base checkpoint grades well below its Polish-adapted descendant at matched architecture, tokenizer, and scale. Mistral-NeMo-12B (base of \pllum{}-12B) scores mean prompt-point $\rhosym=0.102$ against \pllum{}-12B's 0.277, and Llama-3.1-8B (base of \pllum{}-8B) scores 0.193 against \pllum{}-8B's 0.359 -- probe lifts of 0.175 and 0.166 from Polish continual pretraining alone (dispersion lifts 0.094 and 0.223). The Mistral-NeMo base sits at the non-Polish level, so the adaptation, not the architecture, produces the gradation. An exact permutation test over the twelve model-level mean $\rhosym$ values ($\binom{12}{7}=792$ family-label assignments) puts the Polish-vs-non-Polish gap at 0.315 for the probe (Cohen's $d\approx4.1$; $p=0.0013$, the exact null's resolution floor) and 0.179 for dispersion ($p=0.019$); the partial correlation with Polish adaptation controlling for log-size is 0.905. Because checkpoints within a family share corpora, tokenizers, and training pipelines, they are not independent replications: the effective sample is closer to four families than twelve models, so we emphasize the effect sizes over the checkpoint-level $p$-values.

\begin{figure}[t]
\centering
\includegraphics[width=\columnwidth]{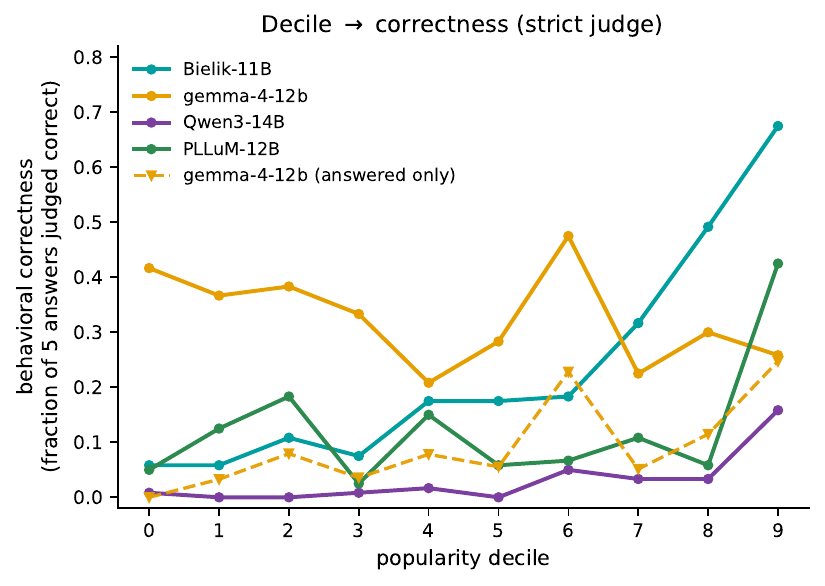}
\caption{Behavioral mirror of the internal gradation: fraction of five sampled answers judged factually correct (strict LLM judge), per popularity decile. \bielik{}-11B rises monotonically 0.06${\,\to\,}$0.68 from decile 0 to 9; \qwen{}-14B confabulates near-floor everywhere; \gemma{}-12B is flat because refusals are judged ``correct''; conditioning on answering restores the gradient (0.00${\,\to\,}$0.25).}
\label{fig:calibration}
\end{figure}

\paragraph{Behavioral mirror.}
We next test whether the internal gradient corresponds to anything behavioral. On the 320-entity evaluation subset per model, we sampled five answers per real entity and judged all of them with a strict factual-correctness rubric (claude-opus-4-8; 4{,}800 calls). Figure~\ref{fig:calibration}: \bielik{}-11B's per-decile correctness rises monotonically from 0.06 (decile 0) to 0.68 (decile 9), mirroring the internal score's gradient. Both correctness and the internal score share popularity as a plausible common cause, so we read the mirror as consistency between the two gradients; whether familiarity predicts correctness beyond pageviews and domain is an incremental-validity question we leave open (\S\ref{sec:limitations}). \qwen{}-14B is near floor everywhere (0.16 even at decile 9): it answers every Polish long-tail question with confident confabulation. \pllum{}-12B is weak except at the very head (0.43 at decile 9). \gemma{}-12B is flat, but confounded: it is the only model that refuses, and the strict judge marks explicit refusals as factually correct ${\sim}88\%$ of the time. Conditioning on \emph{answering}, \gemma{}'s correct-of-non-refused rises from 0.00 (decile 0) to 0.25 (decile 9), noisy but directionally consistent. A second judge outside the Claude family (gpt-5.2, byte-identical rubric) re-judged a stratified 20\% of these verdicts (960 answers): pooled agreement is 89.8\% at the answer level ($\kappa=0.646$) and 91.7\% at the entity level ($\kappa=0.693$), and \bielik{}-11B's decile-correctness monotone survives relabeling (Spearman 0.914 vs.\ 0.906 on the identical subset). The one low-agreement cell (\gemma{}, $\kappa=0.469$) is precisely the refusal-scoring divergence above: the second judge scores refusals as correct far less often (positive rate 0.204 vs.\ Opus's 0.371), corroborating the confound diagnosis. The mirror is not merely a popularity artifact: familiarity retains predictive value for behavioral error after controlling for log-pageviews and domain (likelihood-ratio $p=0.013$ for \bielik{}-11B, $p<0.001$ for \qwen{}-14B), and a calibrated probe gate outperforms a popularity-only gate on \bielik{} (AURC 0.662 vs.\ 0.606).

\section{RQ1: The Signal Is Robust to a Polish/English Stem Substitution}
\label{sec:language}

\begin{figure*}[t]
\centering
\includegraphics[width=\textwidth]{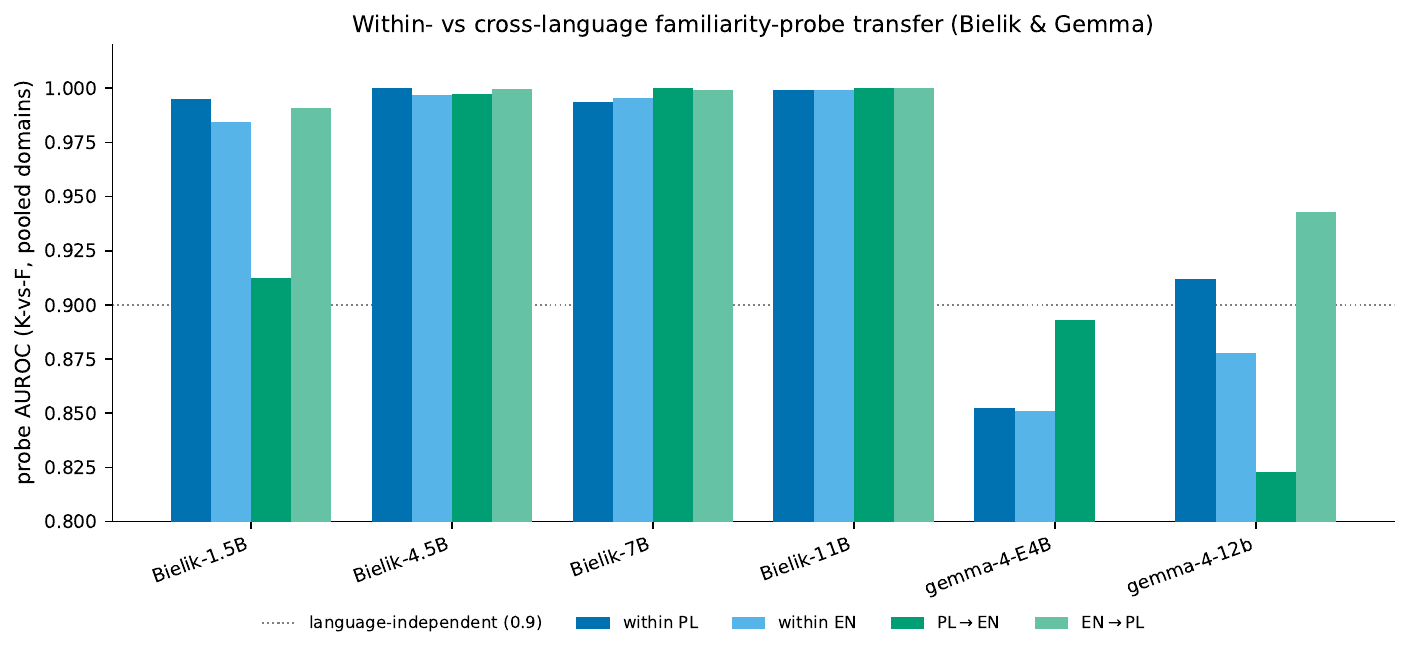}
\caption{Cross-language probe transfer (v1 entities, six models, four domains). A probe trained on Polish questions and applied zero-shot to English questions about the \emph{same} entities (and vice versa) retains 96--101\% of within-language AUROC on known-vs-fabricated (per-model means) and 92--100\% on the fabricated-string-free known-vs-obscure contrast.}
\label{fig:transfer}
\end{figure*}

If the familiarity signal were a property of the question's surface form (Polish morphology, tokenization quirks, template wording), it should degrade when the question language changes. We kept every entity string unchanged and switched only the question stem (\emph{Kim jest \ldots?} $\to$ \emph{Who is \ldots?}), then trained the probe in one language at its CV-best layer and scored the other language zero-shot at the same layer, on identical entities (six models: four \bielik{} scales, two \gemma{} scales; four domains; v1 conditions).

Figure~\ref{fig:transfer}: pooled over domains, cross-language AUROC retains 96--101\% of within-language AUROC on known-vs-fabricated for every model (\bielik{} 0.961--1.005; \gemma{} 0.983--0.985), and 92--100\% on the fabricated-string-free known-vs-obscure contrast. Every model is robust to the stem substitution in this paired setting, consistent with reports of language-agnostic latent processing in multilingual transformers \citep{wendler2024llamas}. Two design limits bound what this shows: the entity name, the largest semantic part of the prompt, is unchanged across conditions, and the cross-language evaluation reuses the entities the probe was trained on in the other language. To test transfer to unseen entities we re-ran the probe with entity-disjoint 5-fold splits (train one language on one entity set, test the other language on held-out entities, both directions). Transfer retains 98--100\% of within-language AUROC on the same held-out split for the four \bielik{} scales, but only 74--93\% for the two \gemma{} scales: robustness to the stem swap generalizes to unseen entities in the Polish-adapted family and degrades for the multilingual model. A single template per language remains a limit (\S\ref{sec:limitations}).

The design also argues against a specific alternative: if \gemma{}'s weaker Polish signal reflected an English-centric model struggling with Polish \emph{questions}, English stems should strengthen its within-language signal. They do not (mean probe delta EN$-$PL: $-0.034$ to $-0.001$ across contrasts and scales). What limits \gemma{} here is its familiarity with the \emph{entities}, not the question language. One caveat is inherent: the language switch also changes the template. A within-Polish neutral-template control in the predecessor study showed the signal is robust to a template swap at these scales, which bounds, but does not eliminate, this confound (\S\ref{sec:limitations}).

\section{RQ3: The Familiarity Direction Is Causally Potent}
\label{sec:causal}

\begin{figure*}[t]
\centering
\includegraphics[width=0.95\textwidth]{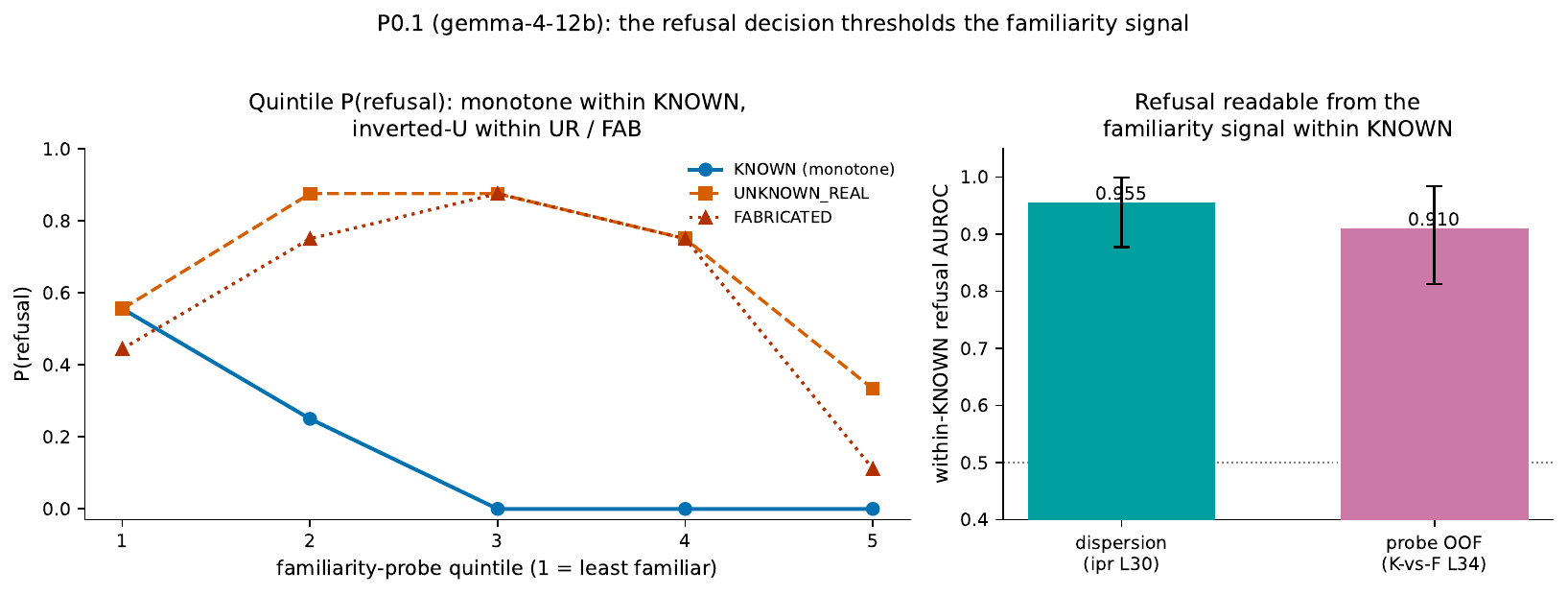}
\caption{Correlational prelude (\gemma{}-12B, v1 athletes): within well-known entities, refusal is readable from the familiarity signal at the independently chosen familiarity layer (dispersion AUROC 0.955), and graded ($\rhosym\approx-0.55$): refusals are concentrated among the known entities assigned lower familiarity. Within unknown/fabricated conditions the mid-band familiarity score does not rank refusals, yet refusal remains strongly decodable from late prompt-point layers, motivating a separate late-layer refusal direction.}
\label{fig:refusalpred}
\end{figure*}

\begin{figure*}[t]
\centering
\includegraphics[width=0.92\textwidth]{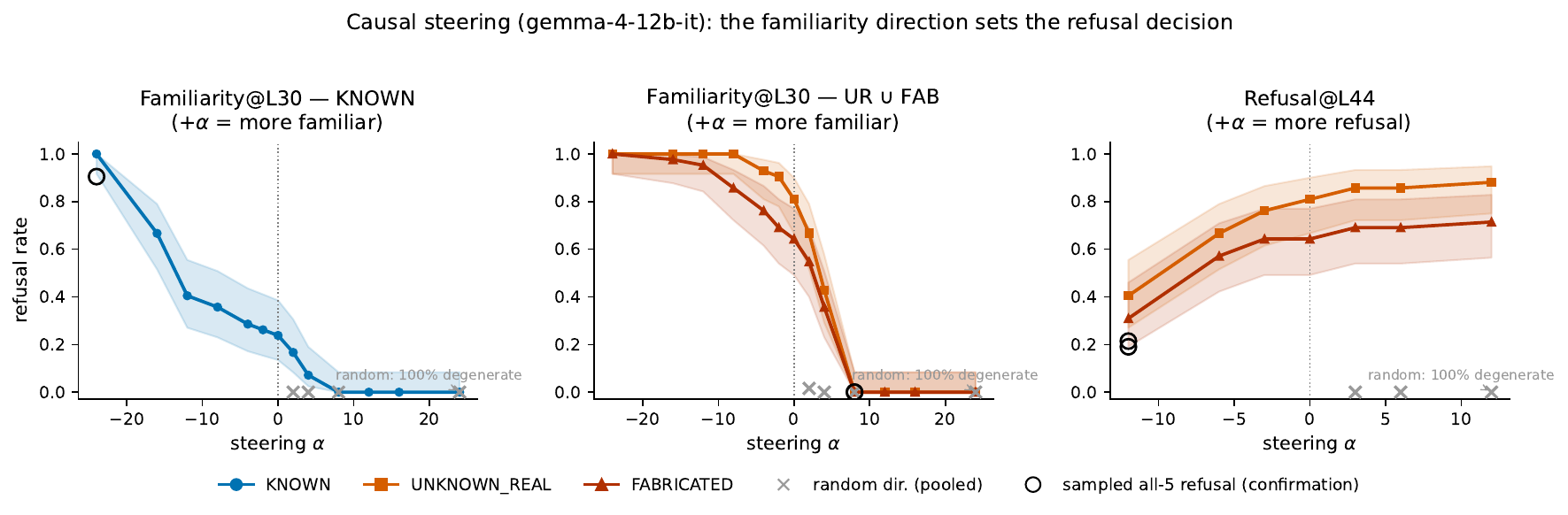}
\caption{Causal dose--response (\gemma{}-12B, rank-one activation addition, single layer, greedy decoding). Steering along the familiarity direction (L30) moves refusal monotonically in both directions: to 1.00 on well-known entities at $\alpha=-24$ and to 0.00 on unknown/fabricated entities at $\alpha\ge+8$; no output triggered our four heuristic degeneration checks at any dose. Equal-norm random directions (pooled, three seeds) never produce a single Polish refusal and instead destroy output coherence. The late refusal direction (L44) is an independent, mostly suppressive handle.}
\label{fig:steering}
\end{figure*}

\begin{table}[t]
\centering
\footnotesize
\setlength{\tabcolsep}{3pt}
\begin{tabular}{lccc}
\toprule
 & \multicolumn{3}{c}{Refusal rate [Wilson 95\% CI]} \\
\cmidrule(lr){2-4}
Condition & $\alpha{=}0$ & fam.\ $\alpha{=}{+}8$ & fam.\ $\alpha{=}{-}24$ \\
\midrule
Known & 0.24 & 0.00 & \textbf{1.00} \\
 & {\scriptsize[.14,.39]} & {\scriptsize[.00,.08]} & {\scriptsize[.92,1.0]} \\
Unknown-real & 0.81 & \textbf{0.00} & --- \\
 & {\scriptsize[.67,.90]} & {\scriptsize[.00,.08]} & \\
Fabricated & 0.64 & \textbf{0.00} & --- \\
 & {\scriptsize[.49,.77]} & {\scriptsize[.00,.08]} & \\
\midrule
\multicolumn{4}{l}{\emph{Refusal direction, L44}\hfill $\alpha{=}0$ \quad $\alpha{=}{-}12$}\\
Fabricated & & 0.64 {\scriptsize[.49,.77]} & 0.31 {\scriptsize[.19,.46]} \\
Unknown-real & & 0.81 {\scriptsize[.67,.90]} & 0.40 {\scriptsize[.27,.56]} \\
\bottomrule
\end{tabular}
\caption{Steering key cells (\gemma{}-12B, athletes, $n{=}42$ per condition). Familiarity steering at L30 saturates refusal in both directions; the late refusal direction at L44 suppresses refusal without touching familiarity. Random equal-norm controls (3 seeds, pooled): refusal 0.00 at every amplitude, but with 100\% of outputs flagged by the degeneration heuristics; no output from the trained directions is flagged.}
\label{tab:steering}
\end{table}

\paragraph{Correlational prelude.}
On \gemma{}-12B, the only model with a refusal behavior to predict, we first asked whether refusal is readable from the familiarity signal (v1 athletes, 42 entities per condition). Within the well-known condition it is: dispersion at the \emph{independently chosen} known-vs-fabricated layer ranks refusals at AUROC 0.955 [0.878, 1.000] (zero-leakage probe 0.910; graded Spearman $\rhosym\approx-0.55$): the seven ``known'' entities \gemma{} refuses are concentrated among those its activations assign lower familiarity. Within the unfamiliar conditions the mid-band score does not rank refusals (AUROC ${\sim}0.6$, CIs spanning 0.5), yet refusal remains strongly decodable from \emph{late} prompt-point layers (out-of-fold 0.96--0.98; raw dispersion 0.89--0.94 at L42--47; Figure~\ref{fig:refusalpred}), suggesting a mid-depth familiarity representation and a late abstention readout downstream of it.

\paragraph{Intervention.}
Correlation alone cannot settle RQ3, so we intervene. We extracted two rank-one directions from prompt-point activations (difference of condition means, unit norm): a \emph{familiarity} direction at layer 30 ($\text{mean(known)} - \text{mean(unknown-real} \cup \text{fabricated)}$) and a \emph{refusal} direction at layer 44 ($\text{mean(refused)} - \text{mean(non-refused)}$ within the unfamiliar conditions). The two directions are geometrically distinct: cosine $-0.14$, against an SD of $0.016$ for random unit pairs in this residual space, a modest negative alignment rather than a shared axis, so the familiarity direction is not a re-labeled refusal direction. During generation we add $\alpha \cdot \mathrm{rms}(h) \cdot d$ to the residual stream at every position of the single target layer, where $\mathrm{rms}(h)=\lVert h\rVert_2/\sqrt{3840}$ keeps doses comparable across layers \citep[cf.][]{turner2023activation,rimsky2024steering,arditi2024refusal}. We ran 5{,}040 greedy generations over a 13-point dose grid, with three equal-norm random directions per layer as controls and a four-way degeneration guard (empty, repetitive, non-Polish, token-salad outputs).

\paragraph{Results (Figure~\ref{fig:steering}, Table~\ref{tab:steering}).}
Pushing well-known entities toward ``unfamiliar'' ($\alpha<0$ along familiarity) raises refusal monotonically from 0.238 to 1.000 at $\alpha=-24$ (dose-curve Spearman $\rhosym=-0.99$); pushing unknown and fabricated entities toward ``familiar'' suppresses refusal from 0.726 to 0.000 at $\alpha\ge+8$ ($\rhosym\le-0.97$). The model then fluently invents biographies: a confabulation proxy (coherent non-refusal answers about unknown entities) rises from 0.27 to 1.00. No steered output triggers any of the four degeneration heuristics; the heuristics screen for gross corruption (empty, repetitive, non-Polish, token-salad outputs) and do not certify preserved semantics or factuality beyond that. Random equal-norm directions never produce a single Polish refusal at any amplitude and instead destroy the output (degeneration 0.39--1.00; at the amplitudes where familiarity steering saturates refusal, random controls are 100\% degenerate), so the effect is not a generic norm artifact. Random directions are a coarse specificity control, however: they show that arbitrary directions are destructive at these amplitudes, not that the familiarity direction is the unique semantic direction with this effect. Five-sample confirmations at temperature 0.7 reproduce the endpoints (known at $\alpha=-24$: mean refusal 0.976, all 42 entities refuse at least once; unfamiliar at $\alpha=+8$: mean refusal 0.019--0.024), and a 60-generation external audit of the refusal marker (claude-haiku-4-5) agrees with the marker metric at 0.95. These results converge with \citet{ferrando2025entity}, who causally modulated refusal via entity-recognition SAE directions in a Gemma~2 chat model; we add the dose--response characterization, equal-norm random controls with degeneration accounting, sampled confirmations, and the late-layer refusal direction below.

\paragraph{The late refusal direction, and a reframing.}
The L44 refusal direction is causally potent but asymmetric: pushing \emph{away} from refusal works (0.726${\,\to\,}$0.357 at $\alpha=-12$; confirmed at five samples), pushing \emph{toward} refusal adds only +0.05 to +0.07 (baseline 0.73--0.81 is near ceiling), and on well-known entities it does nothing. The strict hypothesis motivating it, that the mid-band familiarity direction is causally inert where it fails to \emph{rank} refusals, is falsified: familiarity steering saturates refusal in both directions within the unfamiliar conditions too. Correlational dissociation does not imply causal inertness. The familiarity direction is causally potent for refusal in all three conditions; the late refusal representation adds an independently steerable, mostly suppressive handle. We note the scope of this claim: the intervention is applied at every generation position, not only at the prompt point, and both directions are estimated on the same 42 entities per condition they are evaluated on. A held-out-entity check (correlational, saved activations only) shows the direction is not an artifact of the specific entities: built on a training subset, it separates known from unfamiliar held-out entities at AUROC 0.800 versus 0.802 in-sample, and train-fold directions have cosine 0.999 with the full-set direction. What is demonstrated is that a one-dimensional direction derived from familiarity contrasts strongly steers \gemma{}-12B's refusal behavior; showing that the model's \emph{natural} prompt-time familiarity computation feeds the refusal decision would additionally require prompt-point-only interventions on held-out entities, or mediation-style evidence (\S\ref{sec:limitations}).

\section{RQ4: Gating Generation with a One-Pass Signal}
\label{sec:gating}

\begin{table*}[t]
\centering
\footnotesize
\begin{tabular}{l cccc}
\toprule
Method (cost) & \bielik{}-11B & \gemma{}-12B & \qwen{}-14B & \pllum{}-12B \\
\midrule
\multicolumn{5}{l}{\emph{(a) Real-vs-fabricated entity discrimination}} \\
Our probe (1 pass) & \textbf{0.934} {\scriptsize[.91,.96]} & \textbf{0.871} {\scriptsize[.83,.91]} & \textbf{0.859} {\scriptsize[.81,.91]} & \textbf{0.903} {\scriptsize[.87,.94]} \\
Our dispersion (1 pass) & 0.746 {\scriptsize[.69,.80]} & 0.657 {\scriptsize[.59,.73]} & 0.666 {\scriptsize[.60,.73]} & 0.707 {\scriptsize[.64,.77]} \\
First-token entropy (1 pass) & 0.652 {\scriptsize[.59,.72]} & 0.560 {\scriptsize[.49,.62]} & 0.407 {\scriptsize[.34,.48]} & 0.627 {\scriptsize[.56,.70]} \\
Semantic entropy (5 gen + API) & 0.750 {\scriptsize[.69,.80]} & 0.477 {\scriptsize[.40,.54]} & 0.480 {\scriptsize[.41,.55]} & 0.585 {\scriptsize[.52,.65]} \\
D\textsuperscript{2}HScore (1 gen)\textsuperscript{$a$} & 0.629 {\scriptsize[.55,.69]} & 0.438 {\scriptsize[.36,.51]} & 0.506 {\scriptsize[.44,.58]} & 0.489 {\scriptsize[.42,.56]} \\
EigenTrack (1 gen, sup.)\textsuperscript{$b$} & 0.621 {\scriptsize[.55,.69]} & 0.648 {\scriptsize[.58,.71]} & 0.535 {\scriptsize[.46,.61]} & 0.443 {\scriptsize[.36,.52]} \\
MIND (1 gen, sup.)\textsuperscript{$c$} & 0.653 {\scriptsize[.58,.72]} & 0.677 {\scriptsize[.61,.74]} & 0.717 {\scriptsize[.65,.78]} & 0.601 {\scriptsize[.53,.67]} \\
\midrule
\multicolumn{5}{l}{\emph{(b) Behavioral: majority-incorrect vs.\ correct (strict judge, real entities only)}} \\
Our probe (1 pass) & 0.764 {\scriptsize[.70,.83]} & 0.409 {\scriptsize[.34,.48]} & \textbf{0.914} {\scriptsize[.80,.99]} & 0.722 {\scriptsize[.62,.81]} \\
Our dispersion (1 pass) & 0.675 {\scriptsize[.58,.77]} & 0.262 {\scriptsize[.20,.34]} & 0.866 {\scriptsize[.64,1.0]} & 0.656 {\scriptsize[.52,.78]} \\
First-token entropy (1 pass) & 0.661 {\scriptsize[.57,.75]} & 0.355 {\scriptsize[.29,.43]} & 0.441 {\scriptsize[.24,.67]} & 0.629 {\scriptsize[.51,.74]} \\
Semantic entropy (5 gen + API) & \textbf{0.805} {\scriptsize[.74,.86]} & 0.761 {\scriptsize[.71,.81]} & 0.806 {\scriptsize[.64,.95]} & 0.782 {\scriptsize[.67,.89]} \\
D\textsuperscript{2}HScore (1 gen)\textsuperscript{$a$} & 0.696 {\scriptsize[.62,.77]} & 0.768 {\scriptsize[.70,.83]} & 0.409 {\scriptsize[.21,.61]} & 0.229 {\scriptsize[.13,.33]} \\
EigenTrack (1 gen, sup.)\textsuperscript{$b$} & 0.684 {\scriptsize[.60,.77]} & 0.748 {\scriptsize[.68,.81]} & 0.503 {\scriptsize[.26,.73]} & 0.490 {\scriptsize[.36,.63]} \\
MIND (1 gen, sup.)\textsuperscript{$c$} & 0.528 {\scriptsize[.44,.62]} & \textbf{0.778} {\scriptsize[.72,.84]} & 0.782 {\scriptsize[.52,.98]} & \textbf{0.858} {\scriptsize[.79,.92]} \\
\bottomrule
\end{tabular}
\caption{AUROC (1000-bootstrap 95\% CIs) of one-pass signals vs.\ literature baselines on the v2 evaluation subset ($n{=}320$ per model; contrast (a): 80 fabricated positives; contrast (b): majority-of-5-answers incorrect vs.\ correct under the strict judge, real entities only). Block (b) is heavily imbalanced for \qwen{} (7 majority-correct of 240) and \pllum{} (25 of 240); paired bootstrap: in block (a) the probe's margin over the best baseline excludes zero on every model, in block (b) no best-vs-runner-up difference does, so bold in (b) is numerical, not statistical. Baseline fidelity tiers: \textsuperscript{$a$}D\textsuperscript{2}HScore is training-free and runnable exactly as specified, but its drift component is anti-predictive in our one-sentence-answer regime; \textsuperscript{$b$}EigenTrack instantiates 16 of ${\sim}$22 unspecified spectral features with a window shortened for short answers, trained with 5-fold CV \emph{on eval-pool entities}; \textsuperscript{$c$}MIND's detector (feature + architecture) is trained supervised with CV; its defining unsupervised Wikipedia-continuation data pipeline is not reproduced, so these numbers reflect our adapted variant, not the published method, and the direction of the net bias is unknown. The supervised baselines see 4/5 of the eval pool in training, while our probe never sees any eval entity (trained on disjoint non-eval v2 rows); this supervision asymmetry favors the baselines, but the fidelity gaps above have unknown sign.}
\label{tab:baselines}
\end{table*}

\begin{figure*}[t]
\centering
\includegraphics[width=0.92\textwidth]{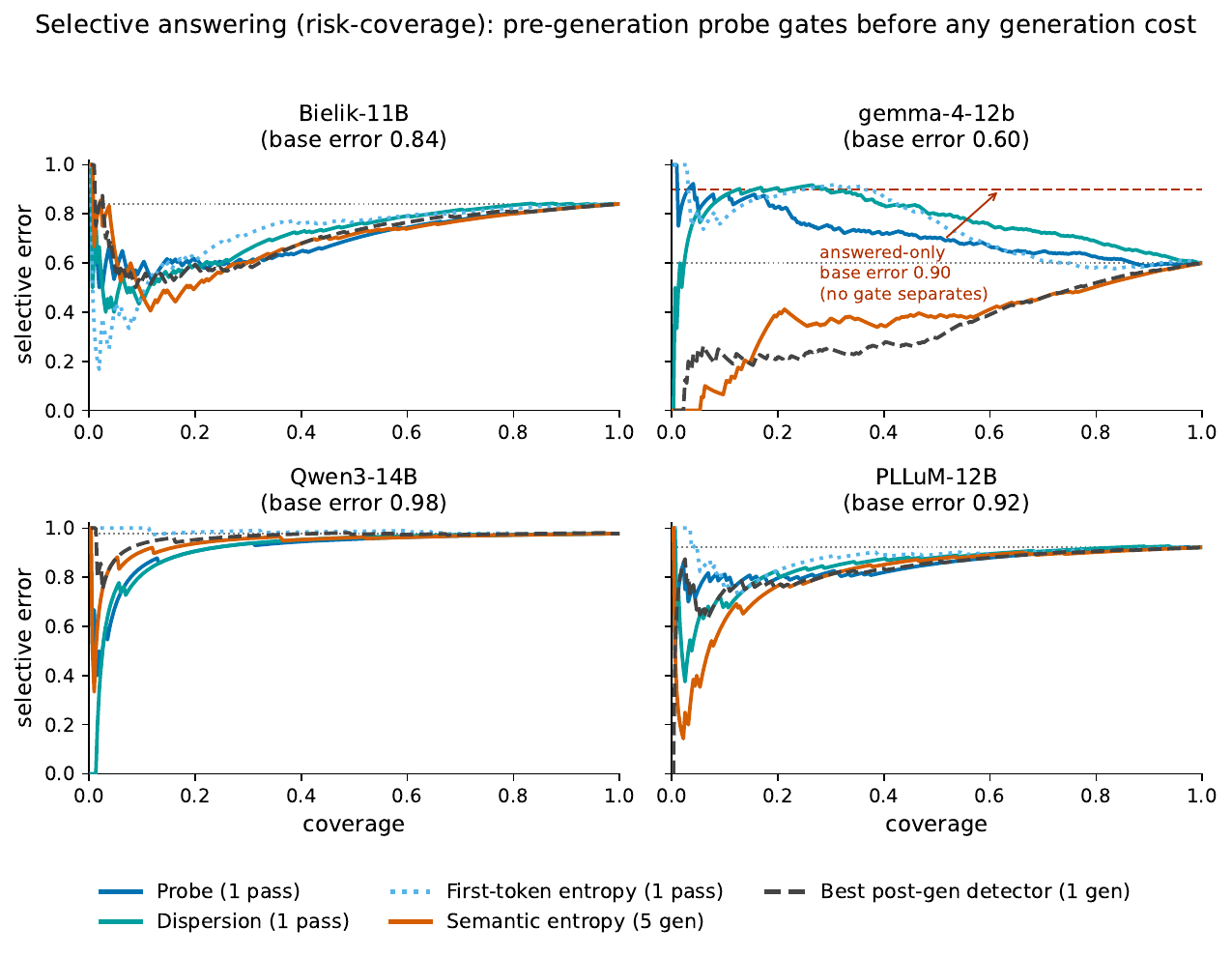}
\caption{Risk--coverage curves for selective answering on the Polish long tail (target: entity answered incorrectly under the strict judge; for \gemma{}, refusals judged ``correct'' lower its apparent base error, \S\ref{sec:limitations}). Base error without gating is severe (0.60--0.98 across models). The calibrated one-pass probe is the best or effectively tied \emph{pre-generation} gate on three of four models; the post-generation detectors pay off only on \gemma{}, i.e.\ only when the model's own generation behavior (refusal) already separates the classes.}
\label{fig:riskcov}
\end{figure*}

\begin{figure}[t]
\centering
\includegraphics[width=\columnwidth]{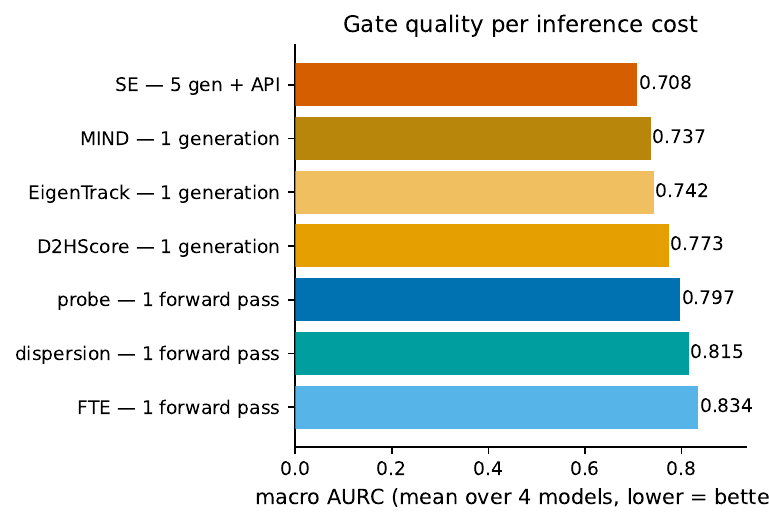}
\caption{Gate quality by inference category (we report pass/generation counts, not measured cost). Semantic entropy attains the best macro AURC (0.708) but requires five sampled generations plus an external clustering call and cannot act before generation; the calibrated probe (macro 0.797) is the strongest gate in the one-forward-pass, pre-generation class, where routing decisions (answer from memory vs.\ retrieve) must be made.}
\label{fig:cost}
\end{figure}

\begin{table}[t]
\centering
\footnotesize
\setlength{\tabcolsep}{1.7pt}
\begin{tabular}{lccccc}
\toprule
Gate (cost) & B-11B & G-12B & Q-14B & P-12B & macro \\
\midrule
SE (5g+API) & 0.697 & 0.381 & 0.944 & 0.811 & 0.708 \\
MIND (1g, sup.) & 0.785 & 0.353 & 0.967 & 0.841 & 0.737 \\
EigenTrack (1g, sup.) & 0.712 & 0.383 & 0.964 & 0.909 & 0.742 \\
D\textsuperscript{2}HScore (1g) & 0.728 & 0.405 & 0.986 & 0.972 & 0.773 \\
\midrule
Probe (1p, pre-gen) & \textbf{0.701} & \textbf{0.712} & 0.923 & \textbf{0.852} & 0.797 \\
Dispersion (1p) & 0.719 & 0.771 & \textbf{0.916} & \textbf{0.852} & 0.815 \\
FT entropy (1p) & 0.722 & 0.745 & 0.985 & 0.886 & 0.834 \\
\midrule
Base error (no gate) & 0.841 & 0.600 & 0.978 & 0.922 & --- \\
\bottomrule
\end{tabular}
\caption{AURC (mean selective error over all coverage prefixes; lower is better; 1g = one generation, 1p = one forward pass, 5g = five sampled generations). \textbf{Bold} marks the best gate \emph{within the pre-generation class} per model, the only class that can route before generating; probe and dispersion tie on \pllum{} at the reported precision (both bolded). All four post-generation detectors have better \emph{macro} AURC (0.708--0.773) than the probe (0.797); their advantage is concentrated on \gemma{}, the sole abstaining model, under a strict-judge target that scores refusals as correct (its 0.60 base error). With refusals removed and risk measured among answered items, \gemma{}'s base error is 0.90 and no gate separates (AURC 0.83--0.93), so that advantage largely disappears (\S\ref{sec:limitations}).}
\label{tab:aurc}
\end{table}

\begin{table}[t]
\centering
\footnotesize
\setlength{\tabcolsep}{3pt}
\begin{tabular}{lcccc}
\toprule
 & \multicolumn{2}{c}{Answers refused} & \multicolumn{2}{c}{Per-answer refusal} \\
\cmidrule(lr){2-3}\cmidrule(lr){4-5}
Model & real & fabr. & people & cities \\
\midrule
\bielik{}-11B & 0/1200 & 0/400 & 0.00 & 0.00 \\
\gemma{}-12B & 533/1200 & 246/400 & 0.52--0.64 & \textbf{0.03} \\
\qwen{}-14B & 0/1200 & 0/400 & 0.00 & 0.00 \\
\pllum{}-12B & 0/1200 & 0/400 & 0.00 & 0.00 \\
\bottomrule
\end{tabular}
\caption{Refusal patterns on the v2 evaluation subset (marker-based). Only \gemma{} abstains, and selectively by \emph{domain type}: 52--64\% per-answer refusal on real people (athletes 0.58, writers 0.64, musicians 0.52) vs.\ 3\% on cities, a biography-caution policy rather than a pure knowledge test. Polish continual pretraining (\pllum{}) did not install any abstention policy: zero marker hits in 1{,}600 answers.}
\label{tab:refusal}
\end{table}

\paragraph{Setup.}
The applied question is whether the one-pass signal can \emph{gate}: decide, before generating, whether to answer from memory, retrieve, or abstain. On the 320-entity evaluation subset per model (six real entities per decile plus 20 fabricated, per domain; stratified, fixed seed) we compare our probe (Platt-calibrated logistic probe on prompt-point residuals, trained only on \emph{non-eval} v2 rows; it never sees an evaluated entity) and dispersion against first-token entropy, discrete semantic entropy over five samples \citep{kuhn2023semantic,farquhar2024detecting}, and three post-generation internal-state detectors: D\textsuperscript{2}HScore \citep{ding2025d2hscore}, EigenTrack \citep{ettori2025eigentrack}, and MIND \citep{su2024mind}, reimplemented from the papers with all adaptations documented in the caption of Table~\ref{tab:baselines} and the released code.

\paragraph{Real-vs-fabricated discrimination.}
On fabricated-vs-real (Table~\ref{tab:baselines}a) the calibrated probe outperforms our adapted implementations of every baseline on every model (0.859--0.934), beating the best generation-time baseline by 0.14--0.30 AUROC per model (paired bootstrap over entities: every per-model margin's 95\% CI excludes zero), despite the supervised baselines training on eval-pool entities while the probe never does. On this target, the prompt point is easier to read than the generated answer's internal states. Purely lexical classifiers on the entity string top out at AUROC 0.786 (character n-grams; single length and token-count features stay $\le$0.61 under all four family tokenizers), so the probe's 0.859--0.934 sits 0.07--0.15 above the lexical-naturalness ceiling and is not explained by surface name form. We stress that the target is operational: it certifies discrimination of our real/fabricated construction, not possession of answer-relevant knowledge; a model can register that an entity exists while knowing no usable facts about it (\S\ref{sec:limitations}).

\paragraph{Behavioral prediction and the \gemma{} inversion.}
On predicting strict-judge incorrectness (Table~\ref{tab:baselines}b) semantic entropy is the most consistent (0.76--0.81 on all four models): it directly measures answer instability. The probe transfers well where behavior follows knowledge (\qwen{} 0.914, \bielik{} 0.764), though no per-cell best-vs-runner-up difference in this contrast is individually significant under a paired bootstrap (e.g., \qwen{} probe-vs-MIND CI $[-0.12, +0.39]$), so we read block~(b) as pattern-level rather than cell-level evidence. On \gemma{} it appears to \emph{invert} (0.409), but this is a label-semantics artifact, not a property of the signal: the strict judge scores explicit refusals as correct, and \gemma{}'s refusals fall mostly on unfamiliar entities. Separating refusal from correctness with the published marker list (which reconciles to Table~\ref{tab:refusal}'s 533/1200) shows the inversion is a labeling artifact, not a signal property: among \emph{answered} items \gemma{}'s base error is 0.90, and only 13 of 149 answered entities are majority-correct, too few to estimate a behavioral AUROC in either direction. The dissociation that matters here is therefore carried by the refusal behavior itself (Table~\ref{tab:refusal}), not by an inverted AUROC: the familiarity signal tracks the entity's status under our operational measure; behavioral error is additionally shaped by the model's policy. More generally, folding refusals into ``correct'' conflates abstention with factual accuracy, and every block~(b) number, and the risk--coverage analysis below, inherits this target definition for the one abstaining model; a clean analysis would separate answered-vs-abstained from correctness-given-answering, which we flag wherever \gemma{} is affected and leave as the principal evaluation revision.

\paragraph{Selective answering.}
Framed as risk--coverage \citep{elyaniv2010foundations,geifman2017selective} (Figure~\ref{fig:riskcov}, Table~\ref{tab:aurc}), base error rates on this long-tail benchmark are severe: 0.84 (\bielik{}-11B), 0.60 (\gemma{}), 0.98 (\qwen{}), 0.92 (\pllum{}); \gemma{}'s 0.60 counts its refusals as correct, and among \emph{answered} items its base error is 0.90 (\S\ref{sec:limitations}). The gates also differ in inference category (Figure~\ref{fig:cost}; we report pass and generation counts rather than measured latency or FLOPs): semantic entropy achieves the best macro AURC (0.708) but requires five sampled generations plus an external clustering call and can only act \emph{after} answering. Every post-generation detector beats the probe's macro AURC (0.708--0.773 vs.\ 0.797). The probe's claim is a within-class one: it is the best or effectively tied gate in the pre-generation class on three of four models (the \pllum{} margin over dispersion sits at the fifth decimal; on \qwen{}, dispersion is better by 0.007), and pre-generation is the only class that can route before generating. The post-generation detectors beat the probe only on \gemma{}, but this advantage is largely a labeling artifact: once refusals are treated as abstentions and risk is measured among answered items, no gate separates on \gemma{} (AURC 0.83--0.93, against a 0.90 answered base error). Calibration: against the behavioral target the probe's raw ECE is 0.41--0.64, a target mismatch (it is calibrated to the real-vs-fabricated target, not answer incorrectness); a one-dimensional out-of-fold Platt recalibration \citep{platt1999probabilistic,guo2017calibration} onto the behavioral target yields ECE 0.039 / 0.063 / 0.000 / 0.049 on \bielik{}/\gemma{}/\qwen{}/\pllum{}. At these base rates a constant predictor attains ECE 0.000 by construction, so low recalibrated ECE certifies that calibration is achievable, not that the score discriminates; ranking quality is carried by the AURC analysis above.

\paragraph{Who acts on the signal?}
Table~\ref{tab:refusal}: only \gemma{} refuses at all, and selectively by domain type: 52--64\% per-answer refusal on real people vs.\ 3\% on cities, a biography-caution policy layered on top of the knowledge signal. \bielik{}, \qwen{}, and \pllum{} produce zero refusals in 1{,}600 answers each: whatever familiarity information their activations carry, their generation policies never convert it into abstention.

\section{Discussion}
\label{sec:discussion}

The four results compose into one claim: \textbf{prompt-point activations carry a graded entity-familiarity readout that is robust to a Polish/English stem substitution, is more strongly associated with Polish adaptation than with scale in this sample, and can strongly steer the one native abstention policy we observe.} Gradation (\S\ref{sec:gradation}) shows the quantity co-varies with Polish adaptation rather than parameter count, observationally. Language transfer (\S\ref{sec:language}) argues against a large question-stem-language effect in the paired setting, consistent with entity-level recognition features \citep{ferrando2025entity}; entity-disjoint transfer remains to be tested. Steering (\S\ref{sec:causal}) shows the familiarity direction is causally potent for refusal, though establishing that the model's natural prompt-time computation flows through it requires mediation-style evidence. The gating experiments (\S\ref{sec:gating}) show the value and the boundary: the signal separates real from fabricated entities better than our adapted post-generation baselines and can act before a single token is generated, but predicting \emph{behavioral} error additionally requires the model's policy, which varies from \gemma{}'s domain-selective caution to the absence of refusal behavior in the other three families. What distinguishes a family that consumes the signal from one that leaves it inert, for instance whether the familiarity direction is coupled into the model's output pathway, is an open mechanistic question we are pursuing in ongoing work.

The dissociation has a deployment corollary for Polish and other medium-resourced languages. Polish-adapted models carry more graded familiarity information but never abstain; \gemma{} abstains but its readout is less graded. A one-forward-pass calibrated probe supplies the missing piece externally: a familiarity-aware gate that routes long-tail queries to retrieval \citep[cf.][]{asai2024selfrag} before a single token is generated. The representational prerequisite for calibrated abstention is thus present in every family we measured; what is missing in three of them is a policy that consumes it. Whether such a policy can be \emph{installed} in a never-refusing model, by activation-level intervention or by the refusal-aware tuning studied for English models \citep{zhang2024rtuning,cheng2024know}, is the natural next question; we leave a systematic comparison to future work.

\section{Limitations}
\label{sec:limitations}

\textbf{Familiarity is an operational construct.} Our primary discrimination target is real-vs-fabricated status, not validated knowledge: a model can register that an entity exists while holding no answer-relevant facts, and the probe's positive training class (top popularity deciles) couples its score to the popularity axis by construction. The fabricated names are token-length matched, but we did not control character n-grams, name-component frequency, or morphological likelihood, so a lexical-naturalness shortcut cannot be fully excluded; non-neural lexical baselines and real-entity-only controls (e.g., post-cutoff entities) are the appropriate follow-up. Lexical baselines bound the shortcut concern: character-n-gram models reach AUROC 0.786, so the probe's 0.859--0.934 is above but not fully independent of surface form. Familiarity also retains predictive value for behavioral error after controlling for log-pageviews and domain (likelihood-ratio $p=0.013$/$p<0.001$ for \bielik{}-11B/\qwen{}-14B), so the behavioral mirror is not purely a popularity common-cause, though popularity remains a strong baseline.

\textbf{Single steering model and domain.} The steering result is established on one model (\gemma{}-12B, the only one with a refusal behavior to steer) and one domain (athletes, $n{=}42$ per condition; Wilson CIs are wide); whether familiarity steering can \emph{install} abstention in the never-refusing families is untested. Both directions are estimated from the entities they are tested on, so effect sizes may be optimistic; the intervention acts at every generation position, so it demonstrates the direction's potency rather than isolating the natural prompt-point pathway; and the familiarity direction contrasts known entities with obscure-real and fabricated ones pooled, so it may encode correlated features (popularity, real/fabricated status, lexical form) alongside familiarity. High-dose specificity rests on the degeneration contrast plus low-dose non-degenerate controls, and the degeneration heuristics do not certify preserved semantics.

\textbf{Judge circularity and the refusal label.} Correctness labels come from a single model family (claude-opus-4-8). A second judge (gpt-5.2, identical rubric) covers a stratified 20\% of the v2 verdicts at $\kappa=0.65$--$0.69$ (\S\ref{sec:gradation}), substantial but not perfect; behavioral numbers inherit the residual uncertainty, and no human audit was performed. The strict judge scores explicit refusals as correct ${\sim}88\%$ of the time, conflating abstention with factual accuracy for \gemma{}; this affects the block~(b) AUROCs, base error rates, risk--coverage/AURC, and behavioral calibration wherever \gemma{} appears. We report a marker-based separation of answered-vs-abstained from correctness-given-answering (\S\ref{sec:gating}): it reconciles to Table~\ref{tab:refusal} and shows \gemma{}'s answered base error is 0.90, not the refusal-inflated 0.60, and that the post-generation gates' \gemma{} advantage is largely a labeling artifact. An LLM three-way re-judge (refuse/correct/incorrect; claude-opus-4-8; all 4{,}800 real-entity answers) confirms this: it agrees with the marker list at 0.99 per answer (\gemma{} 542 refusals vs.\ the marker's 533) and reproduces the answered-only picture (\gemma{} base error 0.88, 14 of 131 answered entities majority-correct).\footnote{Binary and three-way judges agree on 86\% of the 4{,}800 answers (cf.\ the 89.8\% cross-family agreement, \S\ref{sec:gradation}). Disagreements are of two kinds. The binary judge splits the 543 refusals inconsistently (327 scored correct, 216 incorrect), where the three-way judge labels all \texttt{refuse}; this is the source of the correction. The other 142 are borderline factual answers on long-tail entities (133 flip incorrect${\to}$correct, 9 the reverse), a small non-directional shift: the same entity often draws opposite flips on different samples, so per-entity majority labels and \bielik{}'s decile monotone are unchanged. The residual correct/incorrect variation is thus judge noise of the same order as the cross-family disagreement, not a systematic bias.} A human audit remains future work.

\textbf{Baseline fidelity.} Our MIND variant omits its unsupervised data pipeline and is trained supervised with CV (likely flattering it); EigenTrack's unspecified feature set is instantiated as 16 features with a shortened window; D\textsuperscript{2}HScore runs as specified but its drift term is anti-predictive for one-sentence answers. All adaptations are documented and the supervision asymmetry favors the baselines.

\textbf{Subsample sizes.} Gating and behavioral results use 320 entities per model; per-decile correctness estimates average 24 entities; the behavioral contrast is heavily imbalanced for \qwen{} (7 majority-correct of 240) and \pllum{} (25/240).

\textbf{Popularity proxy.} Pageviews proxy pretraining exposure imperfectly: the 12-month window is contemporary while the models' training cutoffs differ and predate it, and pageviews correlate with article length, link density, and recency. The bottom decile contains zero-pageview stubs with no internal gradient; sitelink counts give materially identical results.

\textbf{Family-level inference.} Only four model families are compared, and checkpoints within a family are correlated replications, so the exposure-vs-scale conclusion is sample-limited (\S\ref{sec:gradation}). We add controlled before/after evidence for both \pllum{} architectures: the Mistral-NeMo-12B and Llama-3.1-8B bases grade at $\rhosym=0.102$ and 0.193 versus their descendants' 0.277 and 0.359, isolating Polish continual pretraining from architecture and scale.

\textbf{Refusal metric.} Refusal is detected by a Polish marker list validated by two independent audits (agreement 0.95--1.00), but steering-induced phrasing outside the list would be missed; the metric is untested on other abstention styles.

\textbf{Language-transfer scope.} The EN/PL switch changes both language and template, and only one template per language is tested; a within-Polish neutral-template control bounds the template effect but was not repeated for every v2 model. Entity strings remain Polish in both conditions by design, so EN prompts are code-mixed rather than fully English. We do test entity-disjoint transfer (\S\ref{sec:language}); it holds for \bielik{} but weakens for \gemma{}. Multiple paraphrases, translated entity forms where aliases exist, and additional languages are still needed before any language-independence claim.

\section*{Ethics and Availability Statement}
We use only public information about public entities, plus invented names screened against real-person collisions. Steering shows abstention can be \emph{suppressed} as easily as induced, so deployed abstention policies are not robust to activation-level manipulation; familiarity gating abstains more on long-tail entities, a coverage-disparity risk. Code, the CSV dataset, and provenance artifacts, including pinned model checkpoints and judge identifiers, are available online.\footnote{\url{https://github.com/agentGreg/bielik-entity-familiarity}} The experiments were orchestrated with AI coding assistance. All results were produced and verified by the author.

\bibliography{refs}

\end{document}